\pdfoutput=1

\documentclass[11pt]{article}

\usepackage[preprint]{acl}

\usepackage{times}
\usepackage{latexsym}
\usepackage{graphicx}  
\usepackage{xspace}
\newcommand{\ours}{\textsc{Self-Demos}\xspace}
\newcommand{\data}{OOD-Toolset\xspace}
\usepackage{multirow} 
\usepackage{booktabs} 
\usepackage{siunitx}
\usepackage{subcaption}
\usepackage{enumitem} 
\usepackage{xcolor}
\usepackage{colortbl}

\usepackage[T1]{fontenc}

\usepackage[utf8]{inputenc}

\usepackage{microtype}

\usepackage{inconsolata}
\usepackage{tcolorbox}

\title{\ours: Eliciting Out-of-Demonstration Generalizability \\ in Large Language Models}

\author{  
Wei He$^1$, { }
Shichun Liu$^1$,  { }
Jun Zhao$^1$, { }
Yiwen Ding$^1$,{ }
Yi Lu$^1$,
\\
\textbf{
Zhiheng Xi$^1$,  { }
Tao Gui$^2$\thanks{\ \  Corresponding authors.},  { }
Qi Zhang$^1$\footnotemark[1],  { }
Xuanjing Huang$^1$}
\\
  {$^1$ School of Computer Science, Fudan University} \\
  {$^2$ Institute of Modern Languages and Linguistics, Fudan University} \\
  \texttt{whe23@m.fudan.edu.cn, \{tgui,qz\}@fudan.edu.cn}
}

\begin{document}
\maketitle
\begin{abstract}
Large language models (LLMs) have shown promising abilities of in-context learning (ICL), adapting swiftly to new tasks with only few-shot demonstrations. 
However, current few-shot methods heavily depend on high-quality, query-specific demos, which are often lacking.
When faced with \textit{out-of-demonstration} (OOD\footnote{OOD refers to ``Out-of-Demonstration'' in this paper, not the commonly understood ``Out-of-Distribution''. Similarly, ID stands for ``In-Demonstration''.}) queries, methods that rely on hand-crafted demos or external retrievers might fail.
To bridge the gap between limited demos and OOD queries, we propose \textbf{\ours}, a novel prompting method that elicits the inherent generalizability in LLMs by query-aware demo generation.
The generated demos strategically interpolate between existing demos and the given query, transforming the query from OOD to ID.
To evaluate the effectiveness of our approach, we manually constructed \textbf{\data}, a dataset in the tool-using scenario with over 300 real-world APIs and 1000 instances, each consisting of three tool-use cases as demos and an OOD query.
Thorough experiments on our dataset and two public math benchmarks have shown that our method can outperform state-of-the-art baselines in the OOD setting. 
Moreover, we conduct a range of analyses to validate \ours's generalization and provide more insights.\footnote{Code \& Data: \href{https://github.com/hewei2001/Self-Demos}{https://github.com/hewei2001/Self-Demos}.}

\end{abstract}

\section{Introduction}

\begin{figure}[!ht]
    \includegraphics[width=0.46\textwidth]
    {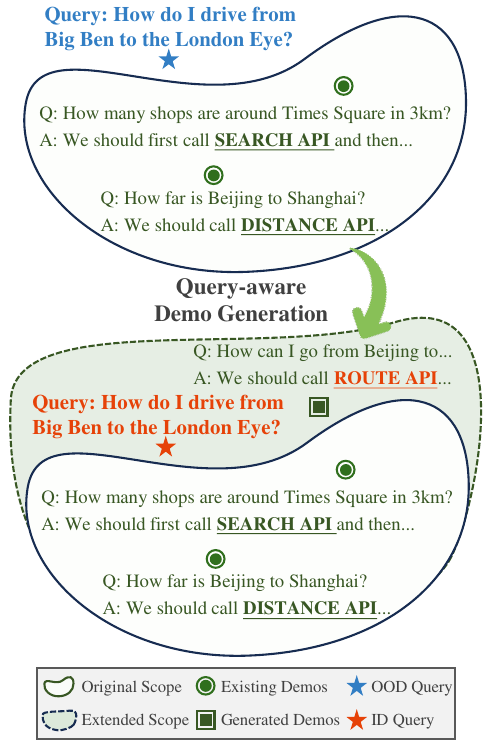} 
    \caption{An example of how query-aware demo generation works.
    In the tool-using scenario, there is a gap between the user query and the available tool-use cases in the original scope since they require different \underline{APIs}. 
    This can lead to errors if the LLM is unfamiliar with the \underline{\textsc{Route} API}.
    After interpolating new demos between the existing ones and the OOD query, LLMs can perform better in the extended scope.}
    \label{fig:scope}
\end{figure}

Large language models (LLMs) have achieved impressive performance across a wide range of tasks, ranging from mathematical reasoning to tool using \cite{brown2022gpt3, kojima2022zero, qin2023tool, xi2023agent}.
The models learn to perform unseen downstream tasks simply by conditioning on a prompt containing input-output pairs (i.e., \textit{few-shot demonstrations}, \citealp{brown2022gpt3}).
This paradigm, also known as in-context learning (ICL), has been found its effectiveness considerably influenced by the quality and relevance of the demos provided \citep{liu2022makes, dong2023survey}.
Thus, how to provide high-quality demos becomes an essential challenge in LLM applications.

The leading few-shot techniques typically hinge on hand-crafted task-specific demos or extensive demo libraries \citep{wei2022cot, liu2022makes, rubin2022learning}. 
However, crafting demos for each unique query is impractical, and the demo libraries are also unable to cover all the potential queries. 
The issue arises when faced with out-of-demonstration (OOD) queries, resulting in poorer performance due to the gap between existing demos and new queries.

An alternative strategy is prompting the LLMs to self-generate relevant demos, thereby guiding themselves toward resolving the query \citep{kim2022sgicl, chen2023self, yasunaga2023analog}.
However, these works often overlook a critical point: instead of blindly recalling relevant demos based on queries, we can perform interpolation between existing demos and queries, as depicted in Figure \ref{fig:scope}.
By strategically interpolating, we can derive more relevant and accurate demos from existing ones, which have proven helpful for the final response \citep{liu2022makes, halawi2023overthinking}.
Specifically, we introduce \textbf{\ours}, a novel prompting method that may fully elicit the model's potential out-of-demonstration generalizability.
Unlike previous works, we developed a complete workflow incorporating pre- and post-processing steps around the demo generation.
Before the demos are generated, we first prompt the model to \textit{``give a general understanding of the user query''}, thereby simplifying the complexity of the analysis in subsequent steps.
Then, we generate query-aware demos and select the most high-quality ones through \textit{Best-of-N sampling} \citep{nakano2021webgpt}.
These selected demos will be used for the final response along with the initial available demos.

To evaluate our approach's efficacy in the OOD context, we manually construct \textbf{\data}, a dataset tailored for tool-using scenarios as delineated by \citet{tang2023toolalpaca}.
Our dataset includes over 300 real-world APIs and 1000 instances, each consisting of three tool-use cases as demos and an OOD query.
Moreover, we benchmarked our method with two public mathematical datasets, GSM8K \citep{cobbe2021gsm8k} and MATH \citep{hendrycks2021math}, to validate its adaptability in different scenarios.
The primary experimental findings reveal that \ours outperforms state-of-the-art baselines in solving OOD queries.
We also conducted ablation studies and other extensive experiments to gain more insights into our method.
Collectively, our analyses show that we have found a more efficient way to elicit the potential OOD generalizability in LLMs.

Our contributions are summarized as follows:
\begin{enumerate}[leftmargin=*]
    \item We proposed \ours, a novel prompting method to elicit the out-of-demonstration (OOD) generalizability in LLMs.
    
    \item We manually constructed \data, a tool-using dataset for better verifying the potential OOD generalizability in LLMs.

    \item We conducted extensive experiments to validate \ours's effectiveness and generalization under different settings.
\end{enumerate}

\section{Related Work}

\subsection{In-Context Learning}
The rise of LLMs such as ChatGPT \citep{chatgpt2022} and LLaMA \citep{touvron2023llama2} has revolutionized the field.
With the model size scaling, LLMs demonstrate remarkable capabilities of ICL \citep{brown2020gpt3, wei2022emergent}, which learns to perform tasks by specific instructions and demonstrations.
Additionally, insights from scaling laws \citep{wei2022emergent} also highlight the LLMs' potential for out-of-distribution generalization.
It refers to the challenge where model inputs deviate from their training distribution \citep{wang2023generalizing}. 
If stimulated effectively, this generalization capability can empower LLMs to address queries outside the training corpus \citep{collins2022structured}, enhancing utility in dynamic and open-ended scenarios.

\subsection{Optimizing Demonstrations for ICL}
The performance of LLMs may be influenced by the quantity, relevance, diversity, and truthfulness of demonstrations \citep{chen2023demonstrations, levy2023diverse, min2022rethinking, halawi2023overthinking}.
There are two primary paradigms to optimize demonstrations and steer models towards generalization.

\paragraph{Demo Retrieval for ICL.}
LLMs are sensitive to the choice of demonstrations.
Therefore, researchers have focused on using retrieval modules to find the most representative demos for ICL. 
One effective strategy is leveraging existing retrievers based on semantic similarity metrics between the available demos and queries \citep{liu2022makes, Agrawal2023selection, gao2023ambiguity, luo2023dricl}.
Another method employs ranking scores derived from fine-tuned language models \citep{rubin2022learning, shi2022xricl}.

\paragraph{Demo Generation for ICL.}
Rather than extracting existing demos, demo generation aims to self-generate exemplars that closely align with the input.
\citet{kim2022sgicl} initially employed language models to produce demos from pre-defined labels. 
Subsequent works adopted a two-stage approach of generating and selecting demos \citep{li2022self, zhang2023autocot, shao2023synthetic}.
In contrast, our work leverages the intrinsic capabilities of LLMs to identify superior demos via best-of-N sampling.

Besides, there are approaches akin to ours.
\citet{chen2023self} adopt multi-steps to construct demonstration pairs, while \citet{yasunaga2023analog} prompt LLMs to recall relevant demos before answering. 
However, our method stands out by combining pre- and post-processing steps around demo generation to guarantee the high quality of generated demos.

\subsection{Eliciting LLMs' Power with Prompts}
Efforts to enhance LLMs include finetuning with specific instructions \citep{wei2022finetuned} and employing prompting strategies like Chain-of-Thought (CoT, \citealp{wei2022cot}).
Our approach adopts the prompt-based strategy and draws inspiration from studies of the ``self'' series \citep{madaan2023sr, wang2023sc, chen2023self}.
The essence of ``self'' is to leverage the model's inherent power, without external modules.
Our method positions the LLM itself as an analyzer, generator, and selector, aiming to elicit its intrinsic generalizability to resolve OOD queries.

\section{Methodology}

\begin{figure*}[!ht]
    \begin{center}
    \includegraphics[width=\textwidth]
    {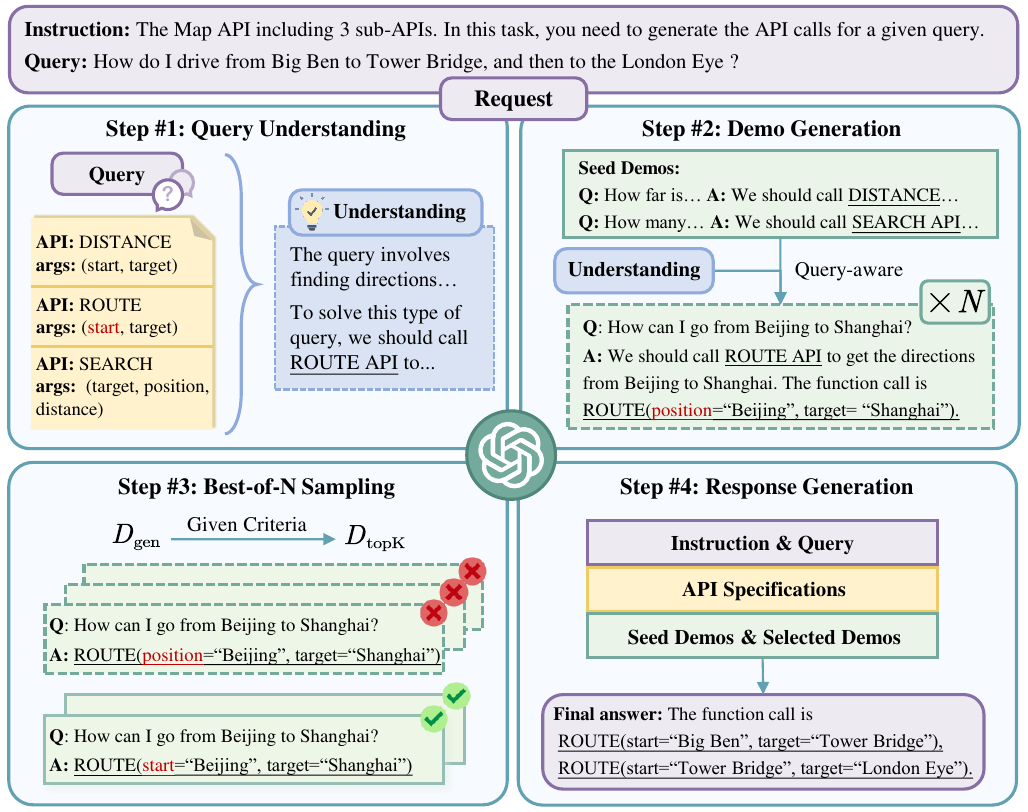} 
    \caption{An overview of the proposed \ours prompting method in tool-using scenario.}
    \label{method}
    \end{center}
\end{figure*}

In this section, we first introduce the construction process of \data. 
Next, we provide a detailed description of the \ours method, which is illustrated in Figure~\ref{method}.

\subsection{\data Construction} \label{sec:construction}

Recent works are evaluated on benchmarks such as BIG-Bench \citep{srivastava2022big} and GSM8K \citep{cobbe2021gsm8k}.
However, since these datasets may have been inadvertently included in the training data of LLMs, there is a risk of overestimating their ability to generalize to OOD query \citep{zhou2023cheater}.
To mitigate this, we chose the tool-using scenarios that are less likely to occur during model training for assessment.
Specifically, we constructed the dataset following the two steps:

\paragraph{Data Collection.}
Our original data derives from the tool-use corpus created by ToolAlpaca \citep{tang2023toolalpaca}.
It was composed of a wide range of real-world APIs complete with API descriptions, usage specifications, and multiple simulated tool-use cases.
However, despite the dataset's comprehensiveness, we noted that the initial AI-generated tool-use cases contain some errors, such as ambiguous queries and incorrect API calls in response.
These minor errors may prevent accurate judgment in our evaluation.
Therefore, we engaged human annotators to manually refine the corpus, producing a high-quality version for more reliable assessment.
Additional details and an example of \data are provided in Appendix \ref{sec:details_of_data}.

\paragraph{OOD Setting.}
We retained the user's queries and corresponding API calls from tool-use cases as input-output pairs for the evaluation.
In addition, we kept the API descriptions and usage specifications from the refined corpus as context for LLMs.
For each test instance, we provided three cases from the same API as initial available few-shot demos (also referred to as \textit{seed demos}, or $D_{\mathrm{seed}}$).
Notably, in the OOD setting, the sub-APIs in seed demos differ from those needed in the final query.

Take the \underline{\textsc{Map}} tool for example, which contains three sub-APIs: \underline{\textsc{Distance}}, \underline{\textsc{Route}}, and \underline{\textsc{Search}} API.
For instance, if the \textsc{Distance} and \textsc{Search} APIs serve as seed demos, the user's query might pertain to the \textsc{Route} API. 
This design tests the model's ability to understand and apply tool-using patterns across different functions, allowing us to explore the OOD generalizability in LLMs.

\subsection{\ours}
We executed the whole workflow by prompting the model itself.
The prompt template for each step is illustrated in Appendix \ref{sec:prompt}.

\paragraph{Query Understanding.}

The first step involves comprehensive query understanding.
Given the model $\mathcal{M}$ and a query $q$, we employ a zero-shot method:
\begin{equation}
u = \mathcal{M}(p_1 \mid\mid q),
\end{equation}
where $p_1$ is the prompt for query understanding, $\mid\mid$ denotes concatenation, and $u$ is the generated understanding.
During this pre-processing step, we aim to reduce the disparity between the initial seed demonstrations and the ultimate target query.
As shown in Figure \ref{method}, when given a query that involves  \underline{\textsc{Map}} API, we guide the model to generate an understanding focused on the more specific \underline{\textsc{Route}} sub-API.
Furthermore, this step resembles a chain-of-thought process \citep{wei2022cot}, which may reduce the cognitive load in subsequent steps.
This is helpful to enhance the relevance and accuracy of the generated demos.

\paragraph{Query-aware Demo Generation.}

Based on the distilled understanding $u$ and seed demos $D_{\mathrm{seed}}$, we generate query-aware demos as:
\begin{equation}
D_{\mathrm{gen}} = \{d_1, d_2,..., d_N\} = \mathcal{M}(p_2 \mid\mid q, u,  D_{\mathrm{seed}}),
\end{equation}
where $p_2$ is the prompt for demo generation, $D_{\mathrm{gen}}$ is the set of generated demos, and $N$ is the number of demos to be generated.
The seed demos, while not directly linked to the specific query, showcase potential tool-using patterns of \underline{\textsc{Map}} API, offering guidance for the generation.
We call the model $N$ times to generate $N$ demos separately, alleviating the difficulty of a single try and avoiding the model falling into consecutive errors in one response.
In this phase, we extend the original scope of the demos to a broader boundary.

\paragraph{Best-of-N Sampling.}

It has been argued that LLMs are unlikely to self-critique their outputs without an external validator \citep{stechly2023gpt4wrong, valmeekam2023can}.
Consequently, we assume that while models might not calibrate and refine outputs, they could still discern the superior output from a variety. 
Therefore, we employ a Best-of-N sampling strategy, where the model is prompted to select the best $K$ demos from the $N$ generated demos based on special criteria:
\begin{equation}
D_{\mathrm{topK}} = \mathcal{M}(p_3 \mid\mid D_{\mathrm{gen}}, C, K),
\end{equation}
where $p_3$ is the prompt for sampling, $D_{\mathrm{topK}}$ is the subset of $K$ demos sampled from the generated ones, conditioned on criteria $C$.

This process is inspired by preference learning, where multiple samples are generated and the one with the highest reward model score is chosen \citep{nakano2021webgpt}. 
It is worth noting that our criteria, which include the demos' accuracy, relevance, and potential helpfulness for the final response, are given to the model via prompts.
Our sampling criteria are more nuanced and do not rely on an external retriever.
This is where \ours differs from methods such as Synthetic Prompting \citep{shao2023synthetic}, which also selects demos after generation.

\paragraph{Response Generation.}

Finally, we leverage the sampled demos $D_\mathrm{topK}$ and the initial seed demos $D_\mathrm{seed}$ to generate the final response:
\begin{equation}
r = \mathcal{M}(p_4 \mid\mid D_{\mathrm{seed}} \cup D_{\mathrm{topK}}, q),
\end{equation}
where $p_4$ is the prompt for response generation, $\cup$ denotes the concatenation of two sets and the $r$ is the final response. 
The concatenation ensures that the model benefits from the query-specific demos in $D_{\mathrm{topK}}$, while also incorporating the beneficial diversity and quality of $D_{\mathrm{seed}}$ \citep{levy2023diverse, halawi2023overthinking}.

\section{Experiments}

To evaluate the effectiveness of \ours, we conduct extensive experiments for comparison and analysis. 

\subsection{Experimental Setups}

\paragraph{Foundation Models.}

We use GPT-3.5 (the \texttt{gpt-3.5-turbo-0613} version) for most of our experiments, with only one additional experiment using the Llama-2-Chat model family, to validate the generalization of \ours across different model sizes.
For all LLMs, we set the parameter $temperature = 0$ for stable responses except for the sampling step, where we set $temperature = 0.7$ to introduce diversity.

\paragraph{Tasks \& Datasets.} \label{sec:dataset}

We evaluate the proposed method in two reasoning-intensive tasks: tool-using and mathematical problem-solving.

In the tool task, we developed the \data for evaluation.
Details of the construction process are described in section \ref{sec:construction}.
In the math task, we employed two public datasets: GSM8K \citep{cobbe2021gsm8k}, featuring elementary math word problems, and MATH \citep{hendrycks2021math}, containing complex problems from high school competitions. 
We evaluate the entire GSM8K testing set and a randomly selected subset from the MATH testing set. 
Distinct OOD settings are designed for math tasks.
For GSM8K, we manually created several outlier samples, ensuring that the testing set did not contain problems with similar contexts.
For MATH, since the problems were categorized into seven subjects and five difficulty levels, we used problems from different subjects but the same level to meet the OOD condition.
The dataset statistics are presented in Table \ref{tab:stat}.

\begin{table*}[!ht]
\centering
\resizebox{0.96\textwidth}{!}{%
\begin{tabular}{cccccc}
\toprule
\textbf{\begin{tabular}[c]{@{}c@{}}Dataset \\ Name\end{tabular}} & \textbf{Size} & \textbf{Demo Source} & \textbf{\begin{tabular}[c]{@{}c@{}}Avg. \#tokens\\ of Query\end{tabular}} & \textbf{\begin{tabular}[c]{@{}c@{}}Avg. \#tokens\\ of Demo\end{tabular}} & \textbf{\begin{tabular}[c]{@{}c@{}}Avg. \#tokens of\\  Context (Few-shot)\end{tabular}} \\ 
\midrule
\textbf{\data} & 1,057 & Same tool, different sub-APIs & 35.5 & 53.8 & 496.0 \\
\textbf{GSM8K} & 1,319 & Manually created outliers & 59.0 & 136.8 & 526.1\\
\textbf{MATH} & 1,000 & Same level, different subjects &69.1  & 291.9 & 1002.1 \\ 
\bottomrule
\end{tabular}%
}
\caption{Statistics of three datasets in the OOD setting.}
\label{tab:stat}
\end{table*}

\paragraph{Evaluation Metric.}

In the report for the math tasks, we present the exact match accuracy for each problem. 
For the tool task, which may require multiple API calls in one case, we assess accuracy using both exact and partial matches. 
Partial matches are awarded half the score if the model's response includes only part of the required API calls.

\subsection{Baselines}
We compare \ours with the following baselines, including two methods that are designed for demo generation:

\paragraph{Zero-shot and Zero-shot + CoT \citep{brown2022gpt3, kojima2022zero}.} 
Prompt the model with the task description, test input, and no demonstration.
Besides, the CoT method integrates a trigger prompt \textit{``let's think step by step''}.

\paragraph{Few-shot \citep{wei2022cot}.} 
Employ a fixed set of seed demos we constructed for each OOD query.
For the GSM8K and MATH datasets, which include solutions with labeled reasoning steps, the demos also feature CoT steps to enhance the model's problem-solving capabilities.

\paragraph{Self-ICL \citep{chen2023self}.}  
A multi-step framework for zero-shot in-context learning by prompting the LLM itself to generate pseudo-inputs and labels.
Unlike our method, they generate inputs and labels separately and then merge them into demos, with no other pre- and post-processing steps.
We have also adapted it into a few-shot variant to make it comparable.

\paragraph{Analogical Prompting \citep{yasunaga2023analog}.}
A single-step prompting method that guides LLM to recall relevant demos and knowledge before solving a given problem.
Here we let it generate demos for the vanilla version and our few-shot variant.
The vanilla Self-ICL and Analogical Prompting methods initially generate three demos each. However, in the few-shot variant, we adjust this to two demos to better align with our approach.

\subsection{Main Results} \label{exp:main}

\begin{table*}[!ht]
\centering
\resizebox{0.885\textwidth}{!}{%
\begin{tabular}{lccccc}
\toprule
\multirow{2}{*}{\textbf{Prompting Method}} & \multicolumn{2}{c}{\textbf{\data}} & \textbf{GSM8K} & \textbf{MATH} & \multirow{2}{*}{\textbf{Average}} \\ 
\cmidrule(lr){2-3}\cmidrule(lr){4-4}\cmidrule(lr){5-5}
& Exact Acc & Part Acc & Acc & Acc &  \\
\midrule
Zero-shot & 64.5 & 68.4 & \,\;75.0$^*$ & \,\;33.0$^*$ & 60.2 \\
Zero-shot + CoT & 66.1 & 70.9 & \,\;75.8$^*$ & \,\;33.9$^*$ & 61.7 \\
Few-shot & 71.9 & 76.6 & 76.2 & 35.1 & 65.0 \\ 
\midrule
Self-ICL (Zero-shot) & 67.0 & 71.1 & 76.6 & 34.6 & 62.3 \\
Self-ICL (Few-shot) & 71.5 & 76.0 & 78.0 & \textbf{37.9} & 65.9 \\
Analogical Prompting (Zero-shot) & 67.8 & 72.0 & \,\;77.8$^*$ & \,\;37.3$^*$ & 63.7 \\
Analogical Prompting (Few-shot) & 71.1 & 75.4 & 75.7 & 36.3 & 64.6 \\
\ours (ours) & \textbf{75.1} & \textbf{79.4} & \textbf{78.2} & \textbf{37.9} & \textbf{67.7} \\ 
\bottomrule
\end{tabular}
}
\caption{Main results of different prompting methods on three datasets. All the results are with GPT-3.5-Turbo. The best performance for each task is in \textbf{bold}. The ($^*$) indicates that results are from \citet{yasunaga2023analog}.}
\label{tab:main_table}
\end{table*}

Table \ref{tab:main_table} shows the performance of each method on three datasets. 
We can find that:
(1) The better performance of few-shot over zero-shot (+ CoT) shows the LLM's capacity to discern and apply underlying patterns from seed demos to OOD queries, indicating a degree of inherent generalizability. 
Furthermore, the \data measures this ability more accurately than the two public math datasets, validating the necessity of creating unseen scenarios and OOD structures of instances.
(2) Only a few-shot method does not fully unlock the model's capability. 
In contrast, the methods with demo generation, especially \ours, present superior performance, underscoring their potential to serve as a reliable prompting strategy in OOD scenarios.
(3) Self-ICL, which generates Q\&A separately, serves a similar purpose to our Best-of-N Sampling step by enhancing the accuracy of generated demos.
Thus, it yields performance that is closest to our method.
However, this framework may also lead to mismatches of Q\&A pairs, i.e., the model fails to answer the questions it generates, which may affect subsequent responses.
(4) Seed demos bring little benefit to the Analogical Prompting method and may even be harmful.
This could be because the additional demos are irrelevant to the instructions of analogical reasoning, which require the model to do multiple tasks.
The seed demos fail to guide the model in different tasks and may distract the model from the whole process.
Overall, \ours outperforms all baselines in solving OOD queries. 

\subsection{Ablation Study} \label{exp:ablation}

\begin{table}[!t]
\centering
\resizebox{0.46\textwidth}{!}{%
\begin{tabular}{lc}
\toprule
\textbf{Pre- \& Post-processing Method} & \textbf{\data} \\ 
\midrule
w/o Pre-processing &  72.9 / 77.5 \\
+ Directly Answering &  72.3 / 77.0 \\
+ Query Understanding & \textbf{75.1} / \textbf{79.4} \\
\midrule
w/o Post-processing & 74.1 / 78.7 \\
+ Self-Critique & 74.3 / 78.8 \\
+ Best-of-N Sampling & \textbf{75.1} / \textbf{79.4} \\
+ Best-of-N Sampling \& Self-Critique & 74.6 / 79.0 \\ 
\bottomrule
\end{tabular}%
}
\caption{Ablation study of pre- \& post-processing methods on \textbf{\data}. The upper rows show the impact of different pre-processing steps, with the other steps remaining consistent with the original. 
The following rows show the impact of post-processing steps, again keeping all other steps consistent with the original.}
\label{tab:ablation}
\end{table}

Table \ref{tab:ablation} presents the results of our ablation study.
We compare a range of pre- and post-processing methods and their influence.

\paragraph{Pre-processing Methods.} \label{exp:pre}

We performed the following settings: no pre-processing before generating demos, directly answering the query before generating, and query understanding before generating.
The result shows that either no pre-processing or directly answering will compromise performance.
Notably, the absence of pre-processing tends to yield homogenous outputs despite our introduction of randomness, potentially due to the model's challenge in reconciling the demanded relevance and diversity. 
Direct answer generation also diminishes performance, as initial errors propagate, leading to more erroneous or ambiguous answers in subsequent steps. 
Hence, a robust pre-processing strategy enhances model performance by ensuring diverse and correct initial responses.

\paragraph{Post-processing Methods.} \label{exp:post}

We performed the following settings: no post-processing after generating two demos, self-critique after generating two, sampling the best two demos after generating five, and self-critique after sampling.
In the self-critique step, we prompt the model to verify and refine the $D_{\mathrm{gen}}$ or $D_{\mathrm{TopK}}$ according to the same criteria $C$.
However, the result indicates that LLMs are no better at verifying their own outputs, echoing the findings of \citet{stechly2023gpt4wrong}.
This also discourages us from constantly improving the quality of demos through iterative verification.

\newcommand{\ddd}[1]{\textcolor{purple}{{#1}}}
\newcommand{\greenuparrow}[1]{\textcolor{green!75!black}{($\uparrow$ #1)}}
\newcommand{\reddownarrow}[1]{\textcolor{red!90!black}{($\downarrow$ #1)}}

\section{Discussion} 

\subsection{Consistency when Model Scaling} \label{exp:scaling}
\begin{figure*}[!ht]
    \begin{center}
    \includegraphics[width=\textwidth]
    {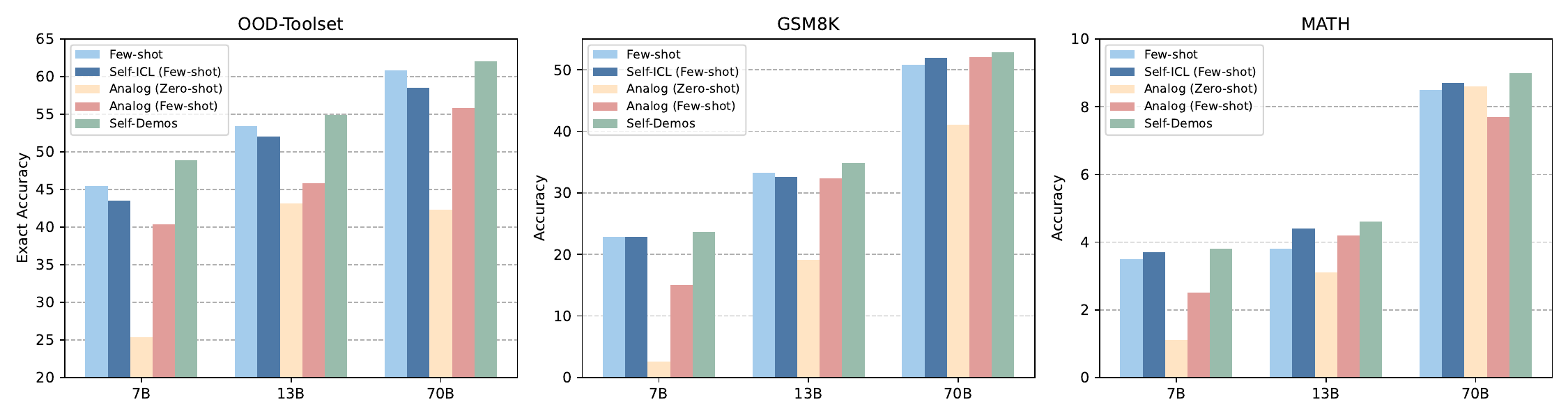} 
    \caption{Performance comparison on Llama-2-Chat model family. \ours consistently improves performance across multiple model sizes from 7B, 13B to 70B parameters.}
    \label{dis:scaling}
    \end{center}
\end{figure*}

Figure \ref{dis:scaling} presents the results on varying sizes of the foundation model, ranging from Llama-2-7B-Chat to Llama-2-70B-Chat.
According to the results, analogical reasoning did not work on smaller models, likely due to their limited capacity to follow hard instructions.
The Self-ICL method encountered similar issues, with the small models' inability to provide accurate demos compromising their effectiveness.
In contrast, our method, which incorporates extra processing steps around demo generation and lowers the task difficulty, proved more adaptable even when the model is weaker ($\sim$10B parameters).
It suggests that our approach is highly adaptable and can be more effective for resource-limited or mobile scenarios.

\subsection{Effectiveness Toward Complex Tasks} \label{exp:complex} 
In the main results, we can observe that both Self-ICL and \ours have shown a considerable improvement on the most challenging MATH datasets. 
This may suggest that the methods of generating demos in advance are more effective for complex tasks, as we will detail here.

\begin{table}[!ht]
\centering
\resizebox{0.46\textwidth}{!}{%
\begin{tabular}{ccccc}
\toprule
\multirow{2}{*}{\textbf{Level}} & \multicolumn{4}{c}{\textbf{Prompting Method}} \\ 
 \cmidrule(lr){2-5}
& FS & Self-ICL + FS & Analog + FS & \ours  \\
 \midrule
1 & 70.2 & 71.3 \greenuparrow{1.6}\,\; & \textbf{80.9} \greenuparrow{15.2} & 74.5 \greenuparrow{6.1}\,\; \\
2 & 58.9 & 61.9 \greenuparrow{5.1}\,\; & \textbf{63.1} \greenuparrow{7.1}\,\; & 58.3 \reddownarrow{1.0}\,\; \\
3 & 37.4 & 38.7 \greenuparrow{3.5}\,\; & \textbf{39.9} \greenuparrow{6.7}\,\; & 39.1 \greenuparrow{4.5}\,\;\\
4 & 28.0 & \textbf{34.7} \greenuparrow{23.9} & 24.0 \reddownarrow{14.3} & \textbf{34.7} \greenuparrow{23.9} \\
5 & 12.4 & 13.8 \greenuparrow{11.3} & 11.6 \reddownarrow{6.4}\,\; & \textbf{14.6} \greenuparrow{17.7}\\ 
\bottomrule
\end{tabular}%
}
\caption{Evaluating prompting methods on the \textbf{MATH} dataset at different complexity levels. The \textbf{Level} corresponds to problem complexity, with higher values indicating greater difficulty. The percentage of performance improvements / declines compared to the few-shot method (FS) is denoted by \textcolor{green!75!black}{($\uparrow$)} / \textcolor{red!90!black}{($\downarrow$)}.}
\label{tab:complexity}
\end{table}

Table \ref{tab:complexity} presents the full results on the MATH dataset across different complexity levels.
Analogical Prompting, as a single-step prompting method, is most effective for simple problems, showing an entirely different trend from the other methods.
This aligns with our previous analysis that high model ability is required for analogical reasoning.
In contrast, Self-ICL and our method significantly gain in more complex problems.
With its greater focus on the relevance and correctness of demos, \ours outperforms others in solving the most difficult level 5 problems.

\subsection{Comparing with Demo Retrieval} \label{exp:retrieval}

A key motivation for our idea is to provide relevant demos for problem-solving, without using an external retriever or demo library.
So, is our approach comparable enough to retrieval-based solutions?
To answer this question, we created two baselines that retrieve exemplars relevant to the given query from external data (i.e., the training set of GSM8K and MATH, which includes labeled Q\&A pairs).
Table \ref{tab:retrieval} shows the results of these methods on two math datasets.

Undoubtedly, the retrieval-based methods perform well, with the dense retriever achieving the highest scores due to its effective representation of latent semantics \citep{karpukhin2020dense}.
Besides, \ours also shows competitive performance, especially on the MATH dataset.
This could be due to more complex questions in the MATH dataset, resulting in intricate semantic connections that cannot be easily captured by a statistical algorithm like BM25 \citep{robertson2009bm25}. 
In contrast, the GSM8K dataset has more uniform and centrally distributed questions, making it more suitable for retrieval-based approaches.

Overall, \ours can still be a good option when resources are limited and retrieval is less feasible.
Moreover, it's worth noting that the techniques of demo generation and retrieval are not mutually exclusive.
Our method is particularly well-suited for a ``\textbf{cold start}'' and once a certain amount of demos is accumulated, we can then employ a complementary retrieval strategy to improve efficiency and reduce incremental costs.

\begin{table}[t]
\centering
\resizebox{0.46\textwidth}{!}{%
\begin{tabular}{lcc}
\toprule
\multirow{2}{*}{\textbf{Demonstrating Method}} & \multicolumn{2}{c}{\textbf{Dataset}} \\
\cmidrule(lr){2-3}
 & \textbf{GSM8K} & \textbf{MATH} \\ 
\midrule
Demo Retrieval (Sparse) & 79.5 & 37.0 \\
Demo Retrieval (Dense) & \textbf{79.7} & \textbf{38.1} \\
Demo Generation (\ours) & 78.2 & 37.9 \\ 
\bottomrule
\end{tabular}%
}
\caption{Comparison with demo retrieval methods on the \textbf{GSM8K} and \textbf{MATH} datasets. The (Sparse) means sparse retrieval using the BM25 algorithm, and the (Dense) means dense retrieval using \texttt{text-embedding-ada-002} API to generate sentence embedding and apply cosine similarity. 
Both baselines retrieve the Top 5 similar samples from the training set as demonstrations.}
\label{tab:retrieval}
\end{table}

\subsection{Number of Demonstrations Matters} \label{exp:number}
\begin{figure}
    \centering
    \begin{subfigure}[b]{0.235\textwidth}
        \centering
        \includegraphics[width=\textwidth]{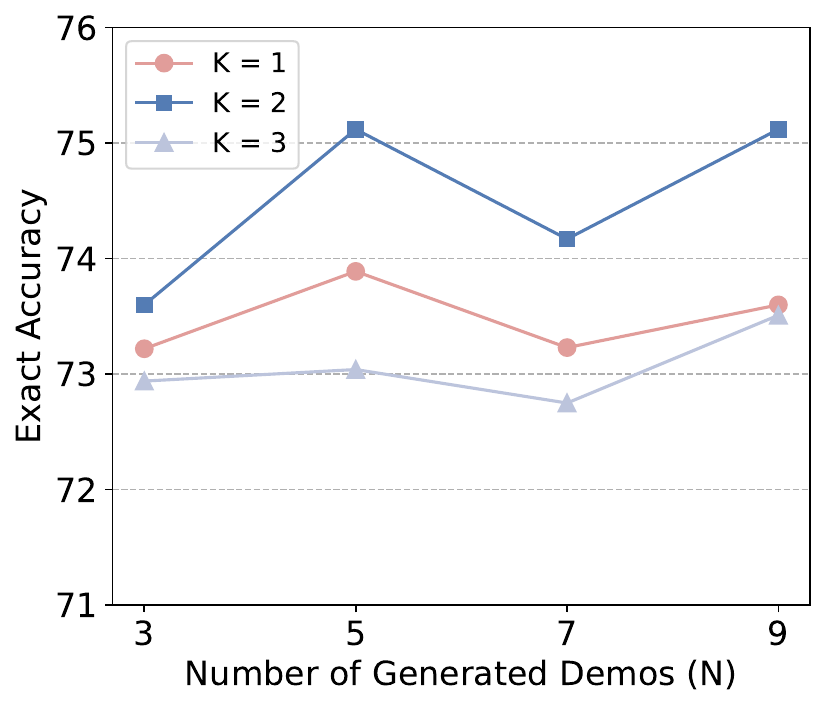}
        \caption{}
        \label{fig:demo_number}
    \end{subfigure}
    \hfill
    \begin{subfigure}[b]{0.24\textwidth}
        \centering
        \includegraphics[width=\textwidth]{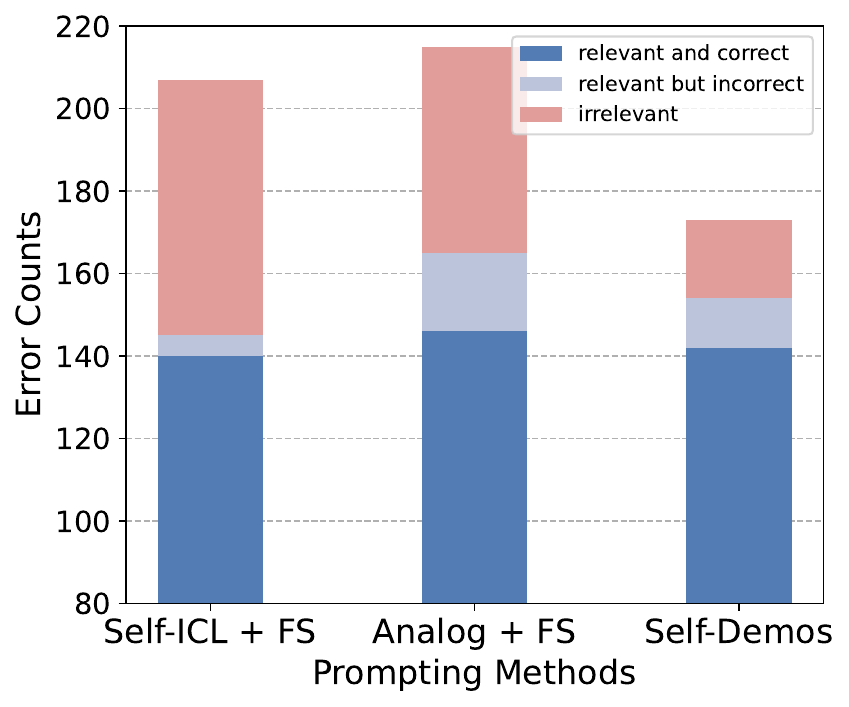}
        \caption{}
        \label{fig:error_analysis}
    \end{subfigure}
    \caption{(a) Comparison of \ours with varying numbers of self-generated demonstrations ($N$) and selected training exemplars ($K$). (b) 
    Error distribution of different methods.
    Demos yielding incorrect answers can be categorized into three types based on relevance and accuracy.
    Both results are on the \textbf{\data}.}
    \label{fig:two_discussion_figures}
\end{figure}

We examine the impact of varying the number of self-generated demos ($N$) and selected demos ($K$) in the tool-using task.
The details are shown in Figure \ref{fig:demo_number}.
Notably, the model performs better when selecting two demos.
We suspect that a singular demo is insufficient to grasp all using patterns of an API and additional samples ($K = 3$) may introduce noise and instabilities and hinder model learning.
Our configuration ($K = 2, N = 5$)  not only maximizes accuracy but also ensures efficiency in computational costs.
In our experiments, we further observed a tendency for the model to preferentially select demos positioned towards the front, indicating the phenomenon of position bias \citep{ko20look, nori23medicine}.

\subsection{Error Analysis} \label{exp:error}

Furthermore, we manually analyze the errors of \ours, comparing with the two baselines of demo generation in Figure \ref{fig:error_analysis}.
Errors were categorized into three distinct types:
(1) \textbf{Irrelevant demos}: These exemplars are generated in a similar distribution and fail to interpolate between seed demos and given queries.
(2) \textbf{Relevant but incorrect demos}: This category includes syntactical errors and redundant or inaccurate parameters. The issues contribute to false information propagation and interfere with the final output.
(3) \textbf{Relevant and correct demos}: Even with correct demonstrations, errors can occur due to the model's inherent limitations and the generalization gap.
Based on Figure \ref{fig:error_analysis}, all three methods have similar results in Category 3 with approximately 140 errors.
However, \ours stands out by greatly lowering the errors in the first two categories.
This suggests that \ours is better at generating relevant exemplars, which improves generalization across novel and unseen tasks.

\subsection{Computational Overhead Analysis} \label{exp:cost}
\begin{table}[t]
\centering
\resizebox{0.46\textwidth}{!}{%
\begin{tabular}{lcc}
\toprule
\textbf{Prompting Method} & \textbf{Cost} & \textbf{OOD-Toolset} \\
\midrule
Few-shot & 0.54 & 71.9 / 76.6 \\
Few-shot + SC (5 Paths) & \underline{2.71} & 72.5 / 77.2 \\
Few-shot + SC (10 Paths) & 5.41 & 72.2 / 77.0 \\
Self-ICL (Few-shot) & \underline{2.37} & 71.5 / 76.0 \\
Analogical Prompting (Few-shot) & 1.21 & 71.1 / 75.4 \\
Self-Demos (Standard) & 4.81 & \textbf{75.1 / 79.4} \\
Self-Demos (KV Cache Reuse) & \underline{2.84} & \textbf{75.1 / 79.4} \\
\bottomrule
\end{tabular}%
}
\caption[Caption for LOF]{Comparison of computational costs on \textbf{OOD-Toolset}. The cost is calculated according to OpenAI price list\protect\footnotemark, measured in dollars per thousand uses. The methods with similar costs are \underline{underlined}.}
\label{tab:cost}
\end{table}

Our method, based on a multi-step framework, naturally leads to additional computational overhead. 
In Table \ref{tab:cost}, we detail this overhead for each method and present another computationally demanding baseline, Self-Consistency, which samples various reasoning paths and generates a consistent answer using a majority vote strategy  \citep{wang2023sc}.
Complete calculation specifics can be found in Appendix \ref{sec:details_of_cost}. 
Statistically, the standard \ours incurs a higher overhead compared to other approaches, primarily due to the demo generation phase that involves repeating the input $N$ times to generate $N$ demos.
\footnotetext{\href{https://openai.com/pricing}{API Pricing - OpenAI API}}
This leads to numerous redundant computations (i.e., KV vectors), a drawback that can be alleviated through caching and reusing \citep{pope2022kvcache}.
It can be achieved by specifying the parameter $n=N$ upon API invocation\footnote{\href{https://platform.openai.com/docs/api-reference/chat/create\#chat-create-n}{API Reference - OpenAI API}}.
The trick cuts overhead by approximately 41\%, reaching computational efficiency on par with Self-ICL and Self-Consistency (5 Paths). 
However, despite Self-ICL's step 2 necessitating multiple calls to model, its distinct query for each input prevents KV cache reuse \citep{chen2023self}.

Moreover, \ours offer substantial \textbf{long-term cost efficiency}. 
When demos are limited, the use of our method does result in a higher computational overhead initially. 
But over time, the high-quality demos that we generate can be preserved, and when a certain amount of them is accumulated, we can apply complementary demo selection methods to reduce the incremental cost and flatten the cost curve.
Refer to Appendix \ref{sec:demo_reuse} for details.

\section{Conclusion}

This paper focuses on addressing the challenge of out-of-demonstration (OOD) queries in few-shot learning scenario.
We present a novel prompting method, \ours, which elicits the OOD generalizability in LLMs by generating query-aware demos.
Our method strategically interpolates between existing demonstrations and the OOD queries, effectively transforming them into in-demonstration (ID) queries.
In an OOD setting, \ours achieved state-of-the-art results on the proposed \data and two public mathematical benchmarks.
For future works, we aim to explore the scalability of the \ours method across diverse domains and to integrate unsupervised learning techniques to refine the quality of generated demos further.

\section*{Limitations}

We summarize the limitations of our method as follows: 
(1) \ours is designed to resolve the out-of-demonstration queries, which can steadily improve downstream task performance, but the process involves additional costs.
In Section \ref{exp:cost}, we explore the computational overhead, allowing users to make informed trade-offs depending on their specific task scenarios.
(2) Our method necessitates certain capabilities of the model.
Although we have done empirical experiments and demonstrated our approach works for weaker models compared to other baselines, it still requires the models to have a certain degree of instruction-following ability.

\section*{Ethics Statement}
In this paper, we have followed ethical standards and principles to ensure the accuracy and validity of our research.
The dataset was manually cleansed to ensure the removal of any sensitive or personal information.
The human-annotated data is collected and used in compliance with relevant ethical guidelines.
During the data construction process, we followed ToolAlpaca's terms under the Apache License 2.0 \citep{tang2023toolalpaca}.

\section*{Acknowledgment}
The authors wish to thank the anonymous reviewers for their helpful comments. This work was partially funded by National Natural Science Foundation of China (No.62206057,61976056,62076069), Shanghai Rising-Star Program (23QA1400200), Natural Science Foundation of Shanghai (23ZR1403500), Program of Shanghai Academic Research Leader under grant 22XD1401100, CCF-Baidu Open Fund, and CCF-Baichuan Fund.

\bibliography{anthology,custom}

\section*{Appendix} \label{sec:appendix}

\renewcommand{\thesubsection}{\Alph{subsection}}
\newcommand{\graytexttt}[1]{\texttt{\textcolor{gray}{#1}}}

\subsection{Supplementary Experiments on GPT-4} \label{sec:demo_reuse}

\begin{table}[!h]
\centering
\resizebox{0.46\textwidth}{!}{%
\begin{tabular}{lcc}
\toprule
\textbf{Model} & \textbf{OOD-Toolset} & \textbf{Cost} \\
\midrule
\rowcolor{gray!10} \multicolumn{3}{c}{\textbf{Few-shot}} \\
GPT-4 & 76.50 / 79.75 & $\sim 1.12$ \\
\midrule
\rowcolor{gray!10} \multicolumn{3}{c}{\textbf{\ours}} \\
GPT-4 in all steps & 80.50 / 83.50 & $\sim 4.95$ \\
GPT-3.5 in all steps & 75.50 / 79.50 & $\sim 0.57$ \\
GPT-3.5 reuse GPT-4 demos in step4 & 76.50 / 79.75 & $\sim 0.13$ \\
\bottomrule
\end{tabular}%
}
\caption{Comparison of performance and overhead on more powerful models (i.e, GPT-4). The cost is calculated according to OpenAI price list, measured by total dollars spent  on 200 instances.}
\label{tab:demo_reuse}
\end{table}

We conducted GPT-4 tests on 200 random OOD-Toolset instances and used its generated demos as inputs for GPT-3.5 in \ours step 4, as detailed in Table \ref{tab:demo_reuse}.

Based on the results, we observe that:
(1) GPT-4's advanced capabilities allow it to match the performance of GPT-3.5 using \ours with simply a few-shot approach. 
However, given the model's enhanced capabilities, it comes with a higher cost.
(2) GPT-4 still benefits from the our proposed method, and the high-quality demos it generates remain effective for weaker models. 
This shows the reusability of demos and proves the way for \ours to reduce long-term costs.

\subsection{Details of \data} \label{sec:details_of_data}
The raw data from ToolAlpaca \citep{tang2023toolalpaca} including the training and testing sets, comprises 468 tool APIs and 4,369 tool-use cases.
Due to a lack of validation of the content generated by GPT-3.5, the dataset may contain specific errors, such as ambiguous queries due to outdated or insufficient information and incorrect API calls due to null or wrong values being passed.
To address these issues, we implemented a data cleansing process in the following steps:

\paragraph{Rule-based Cleaning.}
We structured each tool API in the raw data into a dictionary with keys for \textbf{API Name}, \textbf{Description}, \textbf{Usage Specification}, and \textbf{Tool-use Cases}.
The API name identifies the tool, and the description outlines its purpose.
The usage specification clarifies the API call format and required parameters.
The tool-use cases consist of user queries and corresponding function call lists.
The rule-based cleaning process involved:
\begin{itemize}[leftmargin=*]
\item We removed entries with missing keys and formatting errors, particularly those that did not follow the JSON format in the function calls.
\item We removed user queries that required more than three function calls to be resolved due to their complexity.
\item We removed parameters not directly related to the core functionality of tools, such as API keys and sensitive user information.
\item We removed tools with fewer than 3 instances or fewer than 3 functions to ensure that OOD scenarios could be built.
\end{itemize}

After the first cleaning round, a total of 322 tools and 2,788 instances remained.

\paragraph{Manual Data Cleaning.}
In manual data cleaning, we emphasize the solvability of given queries.
The manual data cleaning process involved:

\begin{itemize}[leftmargin=*]
\item We strive to minimize dependencies between function calls, avoiding scenarios where a subsequent function call relies on the results returned by preceding ones. This is to ensure that these queries can be answered in a round of dialog.
\item While we avoided the exposure of sensitive user information, some necessary parameters within function calls, such as the email address in the email API, are subjected to obfuscation using a placeholder, for instance, \textit{user@example.com}.
\item Time and location information should be explicitly mentioned in the queries, avoiding the use of ambiguous pronouns such as `today', `tomorrow', and `my home'.
\item We confirmed the consistency of parameter values with their data types as defined in the usage specifications.
\end{itemize}

After the second cleaning round, the dataset comprised 321 tools and 2,625 instances.
Table \ref{tab:data_1} presents an illustrative example of the cleaned dataset.

\begin{table*}[!ht]
\centering
\resizebox{0.96\textwidth}{!}{
\begin{tabular}{p{\textwidth}}
\toprule
\textbf{API Name:} \underline{\textsc{Map}}\\
\midrule
\textbf{Description:} \underline{\textsc{Map}} API is used for calculating distances, planning routes, and locating points.\\
\midrule
\textbf{Usage Specifications:}\\
\underline{\textsc{Distance}}: Calculate the distance between two points.\\
Parameters: \{``start'': ``Required. String. The starting point for the distance calculation.'', ``target'': ``Required. String. The destination point for the distance calculation.''\}\\
\\
\underline{\textsc{Route}}: Generate a travel route between two points.\\
Parameters: \{``start'': ``Required. String. The starting point for the route.'', ``target'': ``Required. String. The destination point for the route.''\}\\
\\
\underline{\textsc{Search}}: Locate nearby points within a set distance.\\
Parameters: \{``target'': ``Required. String. The target point to search around.'', ``position'': ``Required. String. The current position of the user.'', ``distance'': ``Required. Integer. The search radius in kilometers.''\}\\
\midrule
\textbf{Tool-use Cases:}\\
Query: How far is Beijing to Shanghai?\\
Function calls: [\underline{\texttt{DISTANCE(start=``Beijing'', target=``Shanghai'')}}]\\
\\
Query: How many shops are around Times Square in 3km?\\
Function calls: [\underline{\texttt{SEARCH(target=``shop'', position=``Times Square'', distance=3)}}]\\
\\
Query: Show me the route from Los Angeles to San Francisco. \\
Function calls: [\underline{\texttt{ROUTE(start=``Los Angeles'', target=``San Francisco'')}}]\\
\\
Query: Are there any bookstores within 5km of Central Park? \\
Function calls: [\underline{\texttt{SEARCH(target=``bookstore'', position=``Central Park'', distance=5)}}]\\
\\
Query: How do I drive from Big Ben to Tower Bridge, and then to the London Eye?\\
Function calls: [\underline{\texttt{ROUTE(start=``Big Ben'', target=``Tower Bridge'')}},\\
\qquad\qquad\qquad\; \underline{\texttt{ROUTE(start=``Tower Bridge'', target=``London Eye'')}}]\\
\\
Query: What’s the distance from my home at 123 Main St to the grocery store at 456 Oak St, and from there to my office at 789 Pine St? \\
Function calls: [\underline{\texttt{DISTANCE(start=``123 Main St'', target=``456 Oak St'')}}, \\
\qquad\qquad\qquad\; \underline{\texttt{DISTANCE(start=``456 Oak St'', target=``789 Pine St'')}}]\\
\bottomrule
\end{tabular}
}
\caption{An illustrative example of the cleaned dataset, composed of four parts: \textbf{API Name}, \textbf{Description}, \textbf{Usage Specifications}, and \textbf{Tool-use Cases}. Among them, the tool-use cases are stored as lists.}
\label{tab:data_1}
\end{table*}

\paragraph{Query and Demonstration Construction.}
After two rounds of data cleaning, the correctness and solvability of the data have been ensured.
Then, we proceeded to select instances from the tool-use cases and construct corresponding demonstrations.
During the selection process, we tended to choose longer instances as queries, considering them to be more challenging.
Following that, we randomly sampled three other instances from the remaining use cases of the same tool as demos.
Note that the sub-APIs to be called for the demos should be different from those required for the chosen queries to fulfill the OOD settings.

Finally, we obtained a set of 1,057 queries, forming our testing set.
Table \ref{tab:data_2} presents an instance of \data.

\begin{table*}[!ht]
\centering
\resizebox{0.96\textwidth}{!}{
\begin{tabular}{p{\textwidth}}
\toprule
\textbf{Seed Demos:} \\
Query: How far is Beijing to Shanghai?\\
Function calls: [\underline{\texttt{DISTANCE(start=``Beijing'', target=``Shanghai'')}}]\\
\\
Query: How many shops are around Times Square in 3km?\\
Function calls: [\underline{\texttt{SEARCH(target=``shop'', position=``Times Square'', distance=3)}}]\\
\\
Query: Are there any bookstores within 5km of Central Park? \\
Function calls: [\underline{\texttt{SEARCH(target=``bookstore'', position=``Central Park'', distance=5)}}]\\
\midrule
\textbf{Query:} How do I drive from Big Ben to Tower Bridge, and then to the London Eye?\\
\bottomrule
\end{tabular}
}
\caption{An instance of \data corresponds to the tool in Table \ref{tab:data_1}, where the function required for the \textbf{Query} is \underline{\textsc{Route}}. Consequently, tool-use cases related to this sub-API should not be included in the \textbf{Seed Demos}.}
\label{tab:data_2}
\end{table*}

\subsection{Prompt Templates} \label{sec:prompt}

The prompt templates of \ours for each step in tool-using tasks are presented in Table \ref{prompt:tool_query_understanding}, \ref{prompt:tool_demo_generation}, \ref{prompt:tool_demo_sampling}, and \ref{prompt:tool_response_generation}.
Similarly, the prompt templates in mathematical problem-solving tasks are presented in Table \ref{prompt:math_query_understanding}, \ref{prompt:math_demo_generation}, \ref{prompt:math_demo_sampling}, and \ref{prompt:math_response_generation}.

\begin{table*}[!ht]
\centering
\resizebox{0.96\textwidth}{!}{
\begin{tabular}{p{\textwidth}}
\toprule
The \graytexttt{\{tool\_name\}} API is used for \graytexttt{\{description\}}. In this task, you need to give a general understanding of the user query and determine which function should be called to solve the query.\\
\\
\textbf{\# Tool Specification:} \\
\graytexttt{\{specification\}}\\
\\
\textbf{\# User Query:} \\
\graytexttt{\{query\}}\\
\\
\textbf{\# Instruction:} \\
Generate a general understanding here. In particular, you need to explicitly indicate the name of the function that should be called.\\
\bottomrule
\end{tabular}
}
\caption{Prompt template for Query Understanding (Step 1) on the \data.}
\label{prompt:tool_query_understanding}
\end{table*}

\begin{table*}[!ht]
\centering
\resizebox{0.96\textwidth}{!}{
\begin{tabular}{p{\textwidth}}
\toprule
The \graytexttt{\{tool\_name\}} API is used for \graytexttt{\{description\}}. In this task, you need to give an example of when to use the API based on the specification.\\
\\
\textbf{\# Tool Specification:} \\
\graytexttt{\{specification\}}\\
\\
\textbf{\# Demonstration:} \\
\graytexttt{\{seed\_demos\}}\\
\\
\textbf{\# Instruction:} \\
Generate an example of how to use the \graytexttt{\{function\_mentioned\_in\_step1\}} function. \\
- After "Query: ", describe the user query.\\
- After "Function Calls: ", give the function calls in the format of ["function\_name(parameter=value)"].\\
\bottomrule
\end{tabular}
}
\caption{Prompt template for Query-aware Demo Generation (Step 2) on the \data.}
\label{prompt:tool_demo_generation}
\end{table*}

\begin{table*}[!ht]
\centering
\resizebox{0.96\textwidth}{!}{
\begin{tabular}{p{\textwidth}}
\toprule
The \graytexttt{\{tool\_name\}} API is used for \graytexttt{\{description\}}. Here are some examples of how to use the API. In this task, you must check the examples for correctness and select one or two best examples to keep. \\
\\
\textbf{\# Tool Specification:} \\
\graytexttt{\{specification\}} \\
\\
\textbf{\# Check List:} \\
- Syntax errors: the function calls should conform to the format like "function\_name(parameter=value)". \\
- Redundant parameters: the function calls must conform to the parameter list in the tool specification. \\
- Value passing errors: the values of parameters should be of the correct type and reasonable.\\
- Unsolvable errors: the query should be solvable with the given function. \\
\\
\textbf{\# Examples to be Checked:} \\
\graytexttt{\{generated\_demos\}} \\
\\
\textbf{\# Instruction:} \\
Select one or two best examples to keep. If there are not enough correct examples, just keep one.\\
For your output: \\
- After "Selection: ", give the serial numbers of your choice in the format of <x>, <y>. \\
- After "Explanation: ", give the reason why you keep the examples. \\
\bottomrule
\end{tabular}
}
\caption{Prompt template for Best-of-N Sampling (Step 3) on the \data.}
\label{prompt:tool_demo_sampling}
\end{table*}

\begin{table*}[!ht]
\centering
\resizebox{0.96\textwidth}{!}{
\begin{tabular}{p{\textwidth}}
\toprule
The \graytexttt{\{tool\_name\}} API is used for \graytexttt{\{description\}}. In this task, you must generate the function calls for a given query.\\
\\
\textbf{\# Tool Specification:} \\
\graytexttt{\{specification\}}\\
\\
\textbf{\# Demonstration:} \\
\graytexttt{\{seed\_demos\}}\\
\graytexttt{\{selected\_demos\}}\\
\\
\textbf{\# Instruction:} \\
Solve the following user query.\\
Query: \graytexttt{\{query\}}\\
Function calls: Give your answer in the format of ["function\_name(parameter=value)"].\\
\bottomrule
\end{tabular}
}
\caption{Prompt template for Response Generation (Step 4) on the \data.}
\label{prompt:tool_response_generation}
\end{table*}

\begin{table*}[!ht]
\centering
\resizebox{0.96\textwidth}{!}{
\begin{tabular}{p{\textwidth}}
\toprule
In this task, you need to give a general understanding of mathematical problems, which can be applied to all similar questions in the same scenario.\\
\\
\textbf{\# Problem:} \\
\graytexttt{\{question\}}\\
\\
\textbf{\# Instruction:} \\
Give a general understanding of this problem in one line. Highlight the general solution methodologies to solve this type of problem. Focus on the problem-solving approach without delving into specific numerical values or answers.
\\
You can refer to this template for your understanding: This problem involves...To solve this type of problem...\\
\bottomrule
\end{tabular}
}
\caption{Prompt template for Query Understanding (Step 1) on the GSM8K and MATH datasets.}
\label{prompt:math_query_understanding}
\end{table*}

\begin{table*}[!ht]
\centering
\resizebox{0.96\textwidth}{!}{
\begin{tabular}{p{\textwidth}}
\toprule
In this task, you need to recall mathematical problems. When presented with a math problem, recall another relevant problem as an example. The example should help answer the initial problem.\\
\\
\textbf{\# Problem:} \\
\textbf{\#\# The initial problem:}\\
\graytexttt{\{question\}}\\
\\
\textbf{\#\# The understanding you can refer to:}\\
\graytexttt{\{understanding\}}\\
\\
\textbf{\# Demonstration:}\\
\graytexttt{\{seed\_demos\}}\\
\\
\textbf{\# Instruction:}\\
Recall one example of a math problem relevant to the initial problem. The example should be distinct from the initial problem (e.g., involving different numbers and names).\\ 
- After "Question: ", describe the problem you generate in one line.\\
- After "Answer: ", explain the step-by-step solution and enclose the ultimate answer in \textbackslash boxed\{\}.\\
\bottomrule
\end{tabular}
}
\caption{Prompt template for Query-aware Demo Generation (Step 2) on the GSM8K and MATH datasets.}
\label{prompt:math_demo_generation}
\end{table*}

\begin{table*}[!ht]
\centering
\resizebox{0.96\textwidth}{!}{
\begin{tabular}{p{\textwidth}}
\toprule
In this task, you need to check the correctness of these math Q\&A pairs and select one or two best examples to keep for answering the initial problem.\\
\\
\textbf{\# The initial problem:} \\
\graytexttt{\{Question\}}\\
\\
\textbf{\# Check List:} \\
- The calculation process in the solution must be correct and without ambiguity.\\
- The examples should be relevant and helpful in solving the initial problem.\\
\\
\textbf{\# Examples to be checked:} \\
\graytexttt{\{generated\_demos\}}\\
\\
\textbf{\# Instruction:} \\
Select one or two best examples to keep. If there are not enough correct and helpful examples, just keep one.\\
For your answer:\\
- After "Selection: ", give the serial numbers of your choice in the format of <x>, <y>.\\
- After "Explanation: ", give the reason why you keep this example.\\
\bottomrule
\end{tabular}
}
\caption{Prompt template for Best-of-N Sampling (Step 3) on the GSM8K and MATH datasets.}
\label{prompt:math_demo_sampling}
\end{table*}

\begin{table*}[!ht]
\centering
\resizebox{0.96\textwidth}{!}{
\begin{tabular}{p{\textwidth}}
\toprule
Your task is to tackle mathematical problems step by step. You can refer to these demonstrations to give your reasoning process.\\
\\
\textbf{\# Demonstration:} \\
\graytexttt{\{seed\_demos\}}\\
\graytexttt{\{selected\_demos\}}\\
\\
\textbf{\# Instruction:} \\
Solve the following problem step by step.\\
Question: \graytexttt{\{Question\}}\\
Answer: Explain the step-by-step solution and enclose the ultimate answer in \textbackslash boxed\{\} here.\\
\bottomrule
\end{tabular}
}
\caption{Prompt template for Response Generation (Step 4) on the GSM8K and MATH datasets.}
\label{prompt:math_response_generation}
\end{table*}


\subsection{Details of Computational Overhead} \label{sec:details_of_cost}

The details about the computational overhead of each methods are shown in Table \ref{tab:detail_of_cost}.

\begin{table*}[!ht]
\centering
\resizebox{0.96\textwidth}{!}{%
\begin{tabular}{lllcc}
\toprule
\textbf{Prompting Method} & \textbf{Avg. \#tokens of Input} & \textbf{Avg.\#tokens of Output} & \textbf{Cost} & \textbf{OOD-Toolset} \\
\midrule
Few-shot & $496.0$ & $22.6$ & 0.54 & 71.9 / 76.6 \\
Few-shot + SC (5 Paths) & $496.0 \times 5=2480.0$ & $22.6 \times 5=113.0$ & \underline{2.71} & 72.5 / 77.2 \\
Few-shot + SC (10 Paths) & $496.0 \times 10=4960.0$ & $22.6 \times 10=226.0$ & 5.41 & 72.2 / 77.0 \\
Self-ICL (Few-shot) & $456.4+498.4\times 2+625.1=2078.3$ & $78.7+23.6\times2+22.2=148.1$ & \underline{2.37} & 71.5 / 76.0 \\
Analogical Prompting (Few-shot) & $598.0$  & $304.5$ & 1.21 & 71.1 / 75.4 \\
Self-Demos (Standard) & $323.6+490.8\times 5+776.4+606.4=4160.4 $ & $3.4+58.0\times 5+7.7+22.5=323.6$ & 4.81 & \textbf{75.1 / 79.4} \\
Self-Demos (KV Cache Reuse) & $323.6+490.8+776.4+606.4=2197.2$ & $3.4+58.0\times 5+7.7+22.5=323.6$ & \underline{2.84} & \textbf{75.1 / 79.4} \\
\bottomrule
\end{tabular}%
}
\caption{Average number of input and output tokens of different methods on \textbf{OOD-Toolset}. In the equation, each term being added represents the average number of tokens per step (used only within a multi-step framework), while each multiplier indicates the number of times that step is called.}
\label{tab:detail_of_cost}
\end{table*}

\subsection{Case Study} \label{sec:case_study}
Even \ours performs better than all other methods, there are instances where it falied while others succeeded. 
We have picked up 3 representative cases for further analysis: (1) \ours succeeded while few-shot / Self-ICL failed, (2) few-shot succeeded while \ours failed, and (3) both failed.
Due to space constraints, we put the full case study in our GitHub repository.

\end{document}